\documentclass[sigplan,twocolumn,nonacm]{acmart}

\renewcommand\footnotetextcopyrightpermission[1]{}

\usepackage{amsmath}
\usepackage{subcaption}
\usepackage{multirow}
\usepackage{pifont}
\usepackage{listings}
\usepackage[linesnumbered,noend,ruled,vlined]{algorithm2e}
\usepackage{xcolor}
\usepackage{graphicx}
\usepackage{color}

\definecolor{ZpfGreen}{RGB}{0,100,0}
\definecolor{ZpfRed}{RGB}{255,0,102}

\newcommand{\stitle}[1]{\vspace*{0.4em}\noindent{\bfseries #1.\/}}

\newcommand{\squishlist}{%
  \begin{list}{$\bullet$}{%
    \setlength{\itemsep}{1pt}%
    \setlength{\parsep}{1pt}%
    \setlength{\topsep}{2.5pt}%
    \setlength{\partopsep}{0.5pt}%
    \setlength{\leftmargin}{1em}%
    \setlength{\labelwidth}{1em}%
    \setlength{\labelsep}{0.6em}%
  }%
}
\newcommand{\squishend}{\end{list}}
\newcommand{\name}{QUILT}

\begin{document}

\title{\name{}: Rethinking Sparse-Attention Prefill through Shared Query Execution}

\author{Zhenduo Zhao}
\authornote{These authors contributed equally to this work.}
\email{zhaozhenduo@huawei.com}
\affiliation{
\institution{Huawei Technologies Co., Ltd.}
\country{China}
}

\author{Qihui Zhou}
\authornotemark[1]
\email{zhouqihui3@huawei.com}
\affiliation{
\institution{Huawei Technologies Co., Ltd.}
\country{China}
}

\author{Mingcong Song}
\authornotemark[1]
\email{songmingcong@huawei.com}
\affiliation{
\institution{Huawei Technologies Co., Ltd.}
\country{China}
}

\author{Zhiyi Chen}
\email{chenzhiyi17@huawei.com}
\affiliation{
\institution{Huawei Technologies Co., Ltd.}
\country{China}
}

\author{Chuangguan Ye}
\email{yechuangguan@huawei.com}
\affiliation{
\institution{Huawei Technologies Co., Ltd.}
\country{China}
}

\author{Fengfan Hou}
\email{houfengfan@huawei.com}
\affiliation{
\institution{Huawei Technologies Co., Ltd.}
\country{China}
}

\author{Zequn Gong}
\email{gongzequn1@huawei.com}
\affiliation{
\institution{Huawei Technologies Co., Ltd.}
\country{China}
}

\author{Jing Li}
\email{joy.lijing@huawei.com}
\affiliation{
\institution{Huawei Technologies Co., Ltd.}
\country{China}
}

\author{Hongjie Si}
\email{sihongjie@huawei.com}
\affiliation{
\institution{Huawei Technologies Co., Ltd.}
\country{China}
}

\author{Guoping Long}
\email{robin3@huawei.com}
\affiliation{
\institution{Huawei Technologies Co., Ltd.}
\country{China}
}

\renewcommand{\shortauthors}{Zhao et al.}

\begin{abstract}
Sparse attention has emerged as a promising approach to reduce the rapidly growing cost of long-context attention. Existing sparse-attention kernels typically adopt query-parallel execution, where different queries independently load, dequantize, and process their selected KV entries. However, such execution is largely workload-agnostic and overlooks the strong correlation among sparse workloads across queries. We observe that neighboring queries frequently select substantially overlapping KV entries. Consequently, processing them independently repeatedly loads and dequantizes the same KV data, resulting in significant redundant memory traffic and computation.

To solve this issue, we present \name{}, a new workload-aware sparse-attention execution mechanism with group-based query execution that explicitly exploits cross-query reuse. \name{} jointly processes neighboring queries and reuses KV entries shared across their sparse workloads, reducing redundant KV data processing and improving hardware utilization. To efficiently realize such execution, \name{} introduces \emph{Shift-and-Compare Set Decomposition (SCSD)}, a new data-parallel algorithm that transforms irregular set operations into regular data-parallel primitives that can be efficiently executed on modern accelerators. \name{} further pipelines SCSD with attention computation to hide its remaining overhead. To accommodate heterogeneous sharing patterns, \name{} is equipped with \emph{cascaded sharing} to capture cross-query reuse hierarchically at multiple granularities. Finally, \name{} adopts a \emph{tile-aware execution} strategy that jointly considers sharing granularity and hardware tile utilization, while selectively removing low-importance query-specific tails to eliminate underutilized tiles.

We evaluate \name{} on LongBench using GLM-5.3 and DeepSeek-3.2 under both tensor and sequence parallelism. Compared with the state-of-the-art sparse-attention kernel, \name{} reduces average kernel latency by up to 55.1\% and the amount of processed KV data by up to 55.9\%, while reducing time-to-first-token (TTFT) latency by up to 36.8\% with negligible accuracy degradation.
\end{abstract}

\maketitle

\section{Introduction}\label{sec:introduction}

\begin{figure}[!t]
	\centering
        \setlength{\abovecaptionskip}{0.1cm}
	\includegraphics[width=0.9\columnwidth]{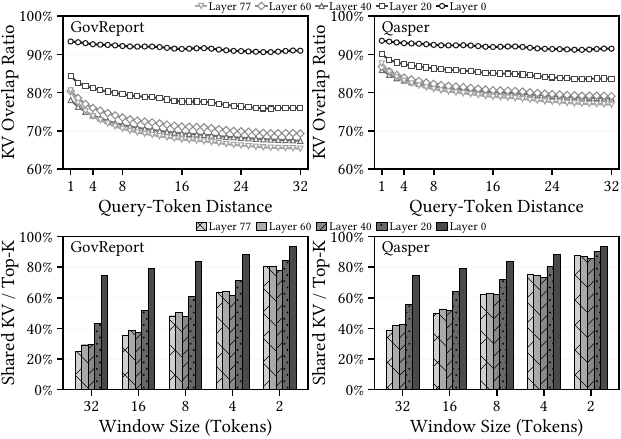}
        \caption{Top-K KV overlap ratios among neighboring queries across model layers and query-group sizes for GLM-5.3~\cite{GLM5} on the Qasper and GovReport datasets in LongBench.}
        \label{fig:intro:overlap}
    \vspace{-5mm}
\end{figure}

The rapid expansion of context windows has enabled large language models (LLMs) to tackle increasingly complex and long-horizon workloads, including long-document understanding~\cite{longbench,leval}, repository-level code analysis~\cite{repobench,swebench}, and agentic applications that maintain extensive interaction histories~\cite{generativeagents,memorybank}. However, supporting such long contexts comes at a substantial inference cost. In particular, during the prefill phase, the model must process the entire input context before generating the first output token, making prefill latency directly visible to users as time-to-first-token (TTFT). As sequence length increases, attention incurs rapidly growing computation and data movement over the expanding key-value (KV) context. Consequently, attention constitutes an increasing fraction of prefill latency at long context lengths, motivating more efficient attention execution for responsive long-context LLM inference.

Sparse attention has emerged as a promising approach to mitigate the growing cost of long-context attention~\cite{arkvale,DeepSeekDSA,FlexPrefill,mInference,quest,h2o,retroinfer,scissorhands,streamingllm,XAttention}. Instead of attending to the entire KV context, each query token employs a lightweight indexer to dynamically identify a small subset of important KV entries and performs attention only over the selected Top-$K$ entries. Since \(K\) is typically much smaller than the context length and can remain bounded as the context grows, sparse attention substantially reduces the computation and memory traffic of long-context prefill. To further reduce the memory footprint and bandwidth cost of KV accesses, sparse attention is often combined with KV quantization, where KV entries are stored in a low-precision format and dequantized on demand when selected for attention computation.

However, sparsity fundamentally changes how attention can be efficiently mapped onto accelerators. In conventional dense attention, multiple queries can be packed together and jointly processed against regular KV tiles on the same compute core, forming sufficiently large matrix operations for efficient execution~\cite{flashattention2,flashinfer}. In dynamic sparse attention, different queries generally select different KV entries, introducing query-specific access patterns that make such efficient multi-query packing largely impractical. Existing sparse-attention kernels~\cite{opstransformer,flashmla} therefore typically adopt a \emph{query-parallel execution strategy}, where different queries are mapped to separate compute cores and independently gather, dequantize, and process their selected KV entries.

While query-parallel execution effectively exploits the massive parallelism of modern accelerators, it treats the KV accesses of different queries independently and overlooks an important workload characteristic: neighboring queries often exhibit highly correlated sparse attention patterns and select substantially overlapping KV entries. To quantify this behavior, we profile GLM-5.3~\cite{GLM5} on LongBench~\cite{longbench} across different datasets, layers, and query-group sizes. As shown in Figure~\ref{fig:intro:overlap}, substantial cross-query overlap consistently exists across model layers and group sizes, and is particularly pronounced in early layers and for smaller query groups. Specifically, when the group size is two, the overlap ratio consistently exceeds 85.91\% and 78.06\% across all layers on Qasper~\cite{qasper} and GovReport~\cite{govreport}, respectively, and further reaches over 90.03\% in the first layer. As the group size increases, the fraction of KV entries shared by all queries in the group gradually decreases, yet considerable overlap persists across most layers. These results reveal substantial opportunities for cross-query KV reuse during prefill, while also showing that the degree of sharing varies across datasets, layers and sharing granularities.

\begin{figure}[!t]
	\centering
        \setlength{\abovecaptionskip}{0.1cm}
	\includegraphics[width=0.9\columnwidth]{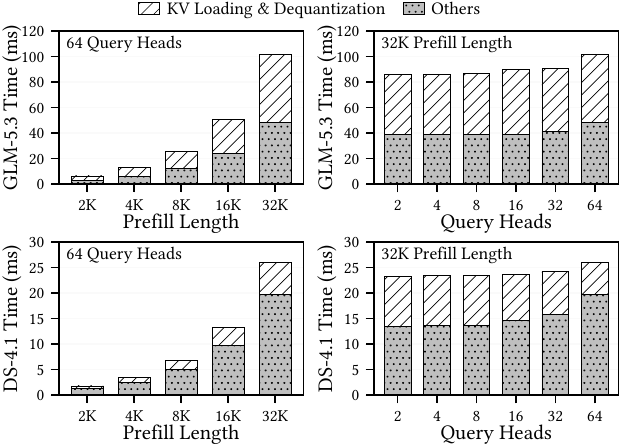}
        \caption{Latency breakdown of the existing sparse-attention kernel for GLM-5.3 and DeepSeek-4.1 on an Ascend 910C NPU across different prefill sequence lengths (SP) and query heads (TP).}
        \label{fig:intro:breakdown}
    \vspace{-6mm}
\end{figure}

Unfortunately, existing query-parallel execution cannot exploit such cross-query reuse. The root cause is that existing sparse-attention kernels are \textbf{workload-agnostic}: they execute each query according to its own Top-$K$ selection without considering the correlation or sharing structure of sparse workloads across queries. Consequently, KV entries shared by neighboring queries are repeatedly loaded from HBM into on-chip memory and independently dequantized on different compute cores, resulting in redundant data movement and computation. As shown in Figure~\ref{fig:intro:breakdown}, our profiling on an Ascend 910C NPU shows that KV loading and dequantization together account for a large portion of the execution time of the existing sparse-attention kernel across different prefill sequence lengths on GLM-5.3~\cite{GLM5} and DeepSeek-4.1~\cite{deepseek41}. This substantial overhead suggests that explicitly exploiting cross-query KV reuse can provide significant opportunities to further improve sparse-attention execution efficiency.

To exploit these cross-query reuse opportunities, we propose \name{}, a new workload-aware sparse-attention execution mechanism with group-based query execution. Instead of processing each query independently, \name{} groups
neighboring queries and jointly executes their correlated sparse workloads
according to the sharing structure of their KV accesses. KV entries shared by multiple queries are loaded from HBM and dequantized only once before being reused across the group, thereby reducing redundant KV data processing. Moreover, jointly processing neighboring queries increases the effective computation granularity of sparse attention and improves accelerator utilization. However, translating such cross-query sharing into actual performance gains is non-trivial and introduces several key design challenges.

Firstly, group-based execution introduces additional overhead for identifying the shared and query-specific KV entries among neighboring queries. A naive implementation based on scalar traversal, explicit loops, or pairwise membership checking can incur substantial set-processing overhead, which may offset or even exceed the benefit of cross-query sharing. To address this challenge, we propose \textbf{Shift-and-Compare Set Decomposition (SCSD)}, a data-parallel algorithm for efficiently computing the intersections and differences of sparse KV sets. SCSD concatenates and sorts the KV indices, transforming irregular set operations into regular shift, comparison, and Boolean primitives that map efficiently onto the parallel execution units of modern accelerators. Moreover, we overlap SCSD with attention computation by embedding it into the existing kernel pipeline, largely hiding its latency behind useful computation. As a result, the kernel can expose cross-query sharing opportunities with near-zero additional overhead on the critical path.

Secondly, the degree of cross-query KV sharing varies substantially across datasets, layers, and sharing granularities, making a fixed query-group size inherently suboptimal. A small group size captures strong local overlap but misses reuse opportunities across groups, whereas a large group size requires KV entries to be shared by more queries and therefore retains only a smaller common subset. To address such heterogeneous sharing patterns, we introduce \textbf{cascaded sharing}, which hierarchically decomposes sparse KV workloads across multiple granularities. Starting from a relatively large query group, the mechanism first extracts KV entries shared by the entire group, then recursively identifies additional sharing within smaller subgroups, and finally processes the remaining query-specific entries. This hierarchical decomposition captures both coarse-grained and fine-grained reuse without committing to a single fixed sharing granularity.

Thirdly, cascaded sharing produces shared and query-specific segments whose sizes are often misaligned with the native attention tile size, causing poorly utilized tail tiles. To address this issue, we introduce \textbf{tile-aware clipping} for the sharing process. For each shared segment, we examine the occupancy of its tail tile. If the occupancy falls below a configurable threshold, we push the tail down to the next cascade level, where it is merged with finer-grained workloads to improve tile utilization. Otherwise, we pad the tail to a full tile and process it at the current sharing level. This push-down is exact and trades a small amount of sharing for improved tile utilization. At the final query-specific level, where no further push-down is possible, we observe that the remaining KV entries are predominantly concentrated near the tail of the original Top-$K$ ranking and thus tend to have lower importance scores. We therefore apply \emph{importance-aware clipping}, pruning the lowest-ranked residual entries to eliminate the final underutilized tile. This approximation further reduces attention-tile execution while introducing negligible accuracy degradation.

We evaluate \name{} on LongBench~\cite{longbench} using GLM-5.3~\cite{GLM5} and DeepSeek-3.2~\cite{DeepSeekDSA} under both tensor and sequence parallelism. Compared with the state-of-the-art sparse-attention kernel, \name{} reduces the average kernel execution latency by up to 55.1\% and the amount of processed KV data by up to 55.9\%. These kernel-level improvements translate into substantial end-to-end benefits, reducing the average time-to-first-token (TTFT) latency by up to 36.8\% while introducing negligible accuracy degradation. These results demonstrate that exploiting cross-query workload structure can effectively translate the inherent correlation in sparse attention into tangible system-level performance gains. To summarize, this paper makes the following contributions:

\squishlist

\item We identify substantial cross-query reuse opportunities in sparse-attention
workloads and reveal a mismatch between their correlated workload
characteristics and existing query-parallel execution, which processes queries
independently and results in redundant KV data processing and inefficient
hardware utilization.

\item We propose \name{}, a new workload-aware sparse-attention mechanism with group-based query execution to exploit cross-query reuse. We develop Shift-and-Compare Set Decomposition for efficient data-parallel set processing and pipeline it with attention computation to hide its overhead. We further introduce cascaded sharing to capture reuse at multiple granularities and tile-aware execution to jointly optimize sharing granularity and hardware utilization.

\item We conduct comprehensive experiments on representative long-context models and workloads under different parallel configurations. Compared with the state-of-the-art sparse-attention implementation, \name{} substantially reduces processed KV data and kernel execution latency, translating these improvements into significant TTFT improvement with negligible accuracy degradation.

\squishend

\section{Background and Motivation}
\subsection{Generative LLM Inference Basics}
\label{subsec:llm-basics}

\stitle{Transformer}
Modern generative LLMs are typically built from stacks of Transformer layers, each consisting of a self-attention module and a feed-forward network (FFN). Given an input sequence $X=[x_1,x_2,\ldots,x_N]$, each layer projects the hidden states into query, key, and value matrices $Q$, $K$, and $V$, and computes
\begin{equation}
S = \frac{QK^T}{\sqrt{d}} + M,\quad
P = \mathrm{softmax}(S),\quad
O = PV,
\notag
\end{equation}
where $M$ is the causal mask and $P_{i,j}$ denotes the attention weight from query token $i$ to key token $j$. The output is then processed by the FFN and passed to the next layer.

\stitle{Prefill and decoding}
LLM inference consists of a \textit{prefill} phase and a \textit{decode} phase. During prefill, all input tokens are processed in parallel to produce the first output token, and the prefill latency directly determines the time-to-first-token (TTFT). Under causal self-attention, each query token attends to the key and value vectors of all preceding tokens. Therefore, for an input sequence of length $N$, dense prefill attention incurs $O(N^2d)$ computation together with rapidly increasing KV data movement as the context grows. After prefill, decoding generates new tokens autoregressively, one token at a time. This work focuses on accelerating the attention computation in the prefill phase.

\subsection{Sparse Attention}
\begin{figure}[!t]
	\centering
        \setlength{\abovecaptionskip}{0.1cm}
	\includegraphics[width=0.85\columnwidth]{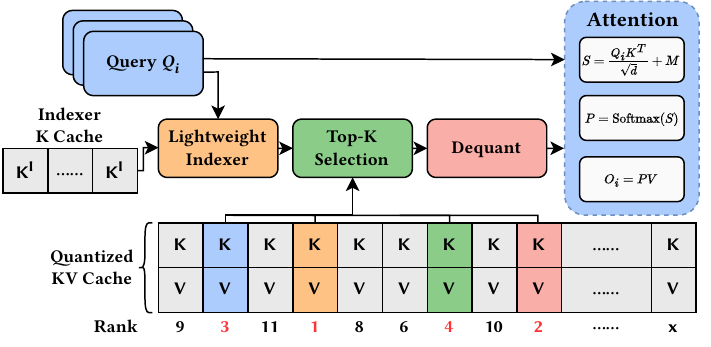}
        \caption{Illustration of the sparse attention workflow.}
        \label{fig:bk:sparse_attention}
    \vspace{-5mm}
\end{figure}

\stitle{The workflow of sparse attention}
Sparse attention reduces the cost of long-context attention by restricting each query token to attend to only a small subset of important KV entries. As shown in Figure~\ref{fig:bk:sparse_attention}, a lightweight indexer first estimates the relevance of previous KV entries to the current query and selects a Top-$K$ subset, where $K$ is typically much smaller than the total context length. Attention is then computed only over the selected KV entries, substantially reducing both computation and memory traffic compared with dense attention. To further reduce memory footprint and bandwidth consumption, the KV cache is often stored in a low-precision format and the selected entries are dequantized on demand before attention computation.

\begin{figure}[!t]
	\centering
        \setlength{\abovecaptionskip}{0.1cm}
	\includegraphics[width=0.95\columnwidth]{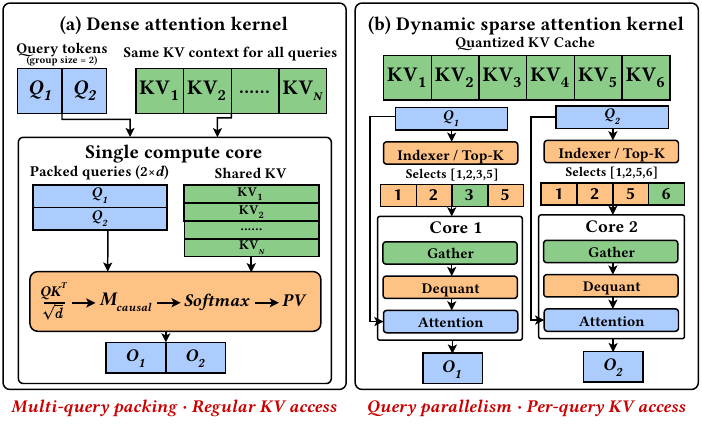}
        \caption{Comparison of the execution workflows of dense and sparse attention kernels.}
        \label{fig:bk:dense_sparse_attention_compare}
    \vspace{-6mm}
\end{figure}

Importantly, sparsification fundamentally changes the execution workflow of the attention kernel. As illustrated in Figure~\ref{fig:bk:dense_sparse_attention_compare}, dense attention operates on regular KV tiles that can be jointly processed by multiple queries, whereas sparse attention introduces query-specific and irregular KV selections. Consequently, sparse kernels typically need to independently gather, dequantize, and process the selected KV entries for each query, leading to an execution structure substantially different from that of dense attention.

\stitle{Training-free sparsification}
Early approaches such as MInference~\cite{mInference}, FlexPrefill~\cite{FlexPrefill}, and XAttention~\cite{XAttention} sparsify pretrained LLMs without modifying their original training procedure. Instead, they exploit the intrinsic structure of attention maps at inference time and use lightweight proxy computations to identify important KV entries. For example, MInference detects recurring structured sparsity patterns and constructs sparse attention indices accordingly, while FlexPrefill adaptively selects sparse patterns and computation budgets based on the observed attention characteristics of individual inputs and heads. XAttention further uses lightweight antidiagonal scoring to estimate block importance and select the most relevant KV blocks for attention.

\stitle{Model-native sparse attention}
More recently, sparse attention has increasingly been incorporated directly into model training and architecture. Native Sparse Attention (NSA)~\cite{deepseeknsa} demonstrates that sparse attention can be trained end-to-end, while DeepSeek-V3.2-Exp~\cite{DeepSeekDSA} further integrates DeepSeek Sparse Attention (DSA) as a native component of a large-scale LLM, using a trainable lightweight indexer to dynamically select important KV entries during both training and inference. Similar designs have since been adopted and extended by recent models, including the GLM-5 series~\cite{GLM5}, DeepSeek-V4 series~\cite{deepseekv4,deepseek41}, and Hunyuan-4 series~\cite{hunyuan4}, suggesting an emerging trend toward model-native sparse attention for efficient long-context processing. 


Importantly, recent models are adopting increasingly sophisticated sparsification strategies. DeepSeek-4.1~\cite{deepseek41}, for example, combines sparse-index generation, refinement over restricted candidate sets, and cross-layer index reuse. Despite such increasingly complex mechanisms, the attention kernel ultimately operates on a query-specific set of selected KV entries. \name{} operates directly on these resulting sparse workloads and is therefore agnostic to how sparse indices are generated, refined, or reused. Consequently, \name{} remains applicable to a broad range of existing and emerging sparse-attention architectures, including both training-free and model-native designs, without requiring changes to their sparsification strategies.

\subsection{Architecture of Modern Accelerators}

\stitle{Hardware components and functions}
Modern AI accelerators typically integrate heterogeneous compute units specialized for different classes of operations. Matrix-intensive operations, such as GEMM and the QK/PV computations in attention, are executed on high-throughput matrix engines, e.g., Tensor Cores on GPUs and Cube units on Ascend NPUs. In contrast, element-wise operations, reductions, comparisons, and other lightweight computations are mainly handled by general-purpose execution units following SIMD/SIMT execution models, such as CUDA cores on GPUs and Vector units on Ascend NPUs. These compute units are backed by a hierarchical memory system, where data are transferred from off-chip HBM to faster on-chip memories before being consumed by the compute units. Importantly, matrix and vector units can execute concurrently when their data dependencies permit, enabling kernels to pipeline heterogeneous computations.

\stitle{Tile-based execution}
Modern accelerator kernels typically execute large tensor operations in a \emph{tile-based} manner. Large tensors are first partitioned into smaller tiles that can be distributed across compute cores and efficiently accommodated by the on-chip memory hierarchy. Within each tile, computation is further mapped to the native execution granularity of the underlying hardware. Matrix engines operate on fixed or constrained matrix fragments, whereas vector units process a fixed number of elements per SIMD/vector instruction. Therefore, the effective tile shape is jointly determined by hardware instruction granularity, data type and alignment requirements, on-chip memory capacity, and available parallelism. Operations on successive tiles can be organized into a software pipeline, allowing data movement, matrix computation, and vector computation to proceed concurrently.


\subsection{Opportunities and Challenges}\label{subsec:motivation}

\stitle{Mismatch between workload and execution}
As discussed in Section~\ref{sec:introduction}, neighboring queries in sparse attention frequently select highly overlapping KV entries. However, existing query-parallel kernels process each query independently and therefore fail to exploit such cross-query reuse. As illustrated in Figure~\ref{fig:bk:dense_sparse_attention_compare}, dense attention can jointly process multiple queries against regular KV tiles, whereas sparse attention typically falls back to query-independent execution because each query selects a different set of KV entries. Consequently, the same KV entries may be repeatedly loaded from HBM into on-chip memory and independently dequantized across different queries, introducing redundant data movement and computation. Moreover, executing each sparse query in isolation often results in narrow and fragmented matrix operations, limiting the utilization of the high-throughput compute units of modern accelerators.

This mismatch between the correlated sparse workload and query-independent execution motivates us to rethink the execution paradigm of sparse-attention kernels. Rather than treating each query as an isolated workload, we advocate a workload-aware execution strategy that explicitly exploits the structural correlation across queries to improve data reuse, reduce redundant computation, and better utilize the heterogeneous compute resources of modern accelerators.

\stitle{Design challenges}
Translating cross-query sharing opportunities into actual kernel-level speedup is non-trivial and requires addressing several key challenges:

\squishlist

\item[\ding{182}] \textit{Set-decomposition overhead}: Exploiting cross-query sharing requires decomposing neighboring Top-$K$ lists into shared and query-specific KV entries. Naive implementations based on scalar traversal, explicit loops, or irregular membership checking can incur substantial overhead, which may offset or even exceed the savings from KV reuse. Moreover, even after accelerating these set operations, executing them as a standalone preprocessing stage still exposes the remaining overhead on the kernel critical path. Efficient cross-query sharing therefore requires both a highly parallel, hardware-friendly set-decomposition algorithm and an execution strategy that overlaps its computation with the original attention pipeline.

\item[\ding{183}] \textit{Heterogeneous sharing granularity}: Cross-query sharing naturally exists at different granularities. Small query groups capture strong local overlap but miss reuse across groups, whereas large groups only capture KV entries shared by all queries and can therefore miss substantial subgroup-level reuse. Efficient execution must exploit multiple sharing granularities rather than committing to a single fixed group size.

\item[\ding{184}] \textit{Tile underutilization}: Reducing the number of logically processed KV entries does not necessarily translate into proportional hardware savings. The shared and query-specific segments produced by group-based execution are often misaligned with the native attention tile size, resulting in poorly utilized tail tiles. Efficient execution must therefore jointly consider sharing granularity and hardware tile utilization.

\squishend

\section{The Design of \name{}}

\begin{figure}[!t]
	\centering
        \setlength{\abovecaptionskip}{0.1cm}
	\includegraphics[width=0.9\columnwidth]{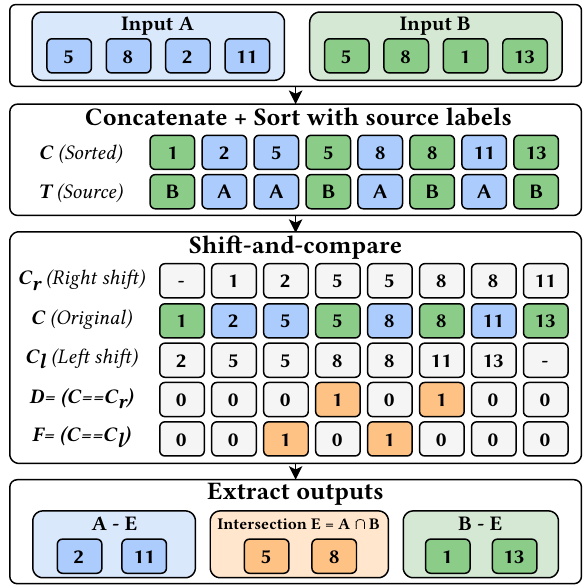}
        \caption{A running example of Shift-and-Compare Set Decomposition (SCSD) with two input KV index sets.}
        \label{fig:design:scsd}
    \vspace{-2mm}
\end{figure}

To address the workload execution mismatch identified in Section~\ref{subsec:motivation}, we propose \name{}, which redesigns sparse-attention execution around cross-query workload sharing. Instead of processing each query independently, \name{} groups neighboring queries and jointly executes their correlated sparse workloads, enabling shared KV entries to be reused across queries while improving accelerator utilization.

Realizing such workload-aware execution mechanism requires addressing the three challenges discussed above. For \textbf{\textit{Challenge 1}}, we propose \emph{Shift-and-Compare Set Decomposition (SCSD)} to transform irregular set operations into hardware-friendly vector operations, and further develop \emph{pipelined SCSD execution} to overlap the remaining set-processing overhead with the original attention pipeline. For \textbf{\textit{Challenge 2}}, we introduce \emph{cascaded sharing}, which hierarchically captures cross-query reuse at multiple sharing granularities instead of relying on a fixed query-group size. For \textbf{\textit{Challenge 3}}, we develop \emph{tile-aware cascaded execution} to jointly optimize sharing granularity and hardware tile utilization, and apply importance-aware clipping to eliminate poorly utilized query-specific tail tiles with negligible accuracy impact.

\subsection{Shift-and-Compare Set Decomposition}

Group-based query execution requires efficiently decomposing the Top-$K$ selections of neighboring queries into shared and query-specific KV entries. A straightforward implementation may rely on scalar merge, explicit loops, or pairwise membership checking between Top-$K$ lists. However, sparse attention already performs a relatively small amount of computation per query, leaving little room for additional bookkeeping overhead. Consequently, a naive set-decomposition implementation can introduce substantial serialized control flow and irregular comparisons, whose cost may offset or even exceed the savings from eliminating redundant KV loading and dequantization. Efficient set decomposition is therefore critical to realizing the benefits of cross-query sharing. To address this challenge, we propose \emph{Shift-and-Compare Set Decomposition (SCSD)}, which transforms irregular set operations into regular sorting, shifting, and equality-comparison primitives that can be efficiently parallelized on modern accelerators.

Figure~\ref{fig:design:scsd} shows a running example of SCSD with two Top-$K$ index lists, $A$ and $B$, each containing $N$ unique KV indices. We use the two-way case for clarity, while the same principle naturally generalizes to multi-way set decomposition. SCSD first concatenates the two input lists into a $2N$-element sequence and jointly sorts the indices while tracking the source of each element:
\begin{equation}
(C,T) = \mathrm{Sort}(A \Vert B),
\end{equation}
where $C$ denotes the sorted indices and $T_i \in \{0,1\}$ indicates whether $C_i$ originates from $A$ or $B$. Since each Top-$K$ list contains unique indices, an index appearing in both $A$ and $B$ occurs exactly twice in $C$. After sorting, these two occurrences are guaranteed to be adjacent.

SCSD therefore identifies shared entries through bidirectional shift-and-compare operations. Specifically, we construct the right- and left-shifted sequences
$C_r$ and $C_l$, and compute
\begin{equation}
D = (C = C_r)
\end{equation}
\begin{equation}
F = (C = C_l)
\end{equation}
For each shared index, the two adjacent copies are captured by $D$ and $F$,
respectively. Selecting one copy using $D$ therefore directly yields the
intersection:
\begin{equation}
E = C[D] = A \cap B
\end{equation}
An index that matches neither of its neighbors must appear in only one input
list. We thus identify all query-specific entries using
\begin{equation}
G = \neg(D \lor F)
\end{equation}
and further separate them according to their source labels:
\begin{equation}
A-E = C[G \land (T=0)]
\end{equation}
\begin{equation}
B-E = C[G \land (T=1)]
\end{equation}

This formulation is particularly well suited to modern accelerators. SCSD
consists of a fixed sequence of regular data-parallel primitives, including
shifts, element-wise comparisons, Boolean operations, and masked selections,
which can be executed concurrently across the Top-$K$ sequence. Unlike
conventional set-intersection implementations based on explicit loops, scalar
pointer advancement, or branch-heavy membership checks, SCSD avoids serialized
scalar execution and irregular control flow. As a result, it exposes abundant
data-level parallelism and substantially reduces the bookkeeping overhead of
cross-query sharing.

\subsection{Pipelined Execution}

\begin{figure}[!t]
	\centering
        \setlength{\abovecaptionskip}{0.1cm}
	\includegraphics[width=1\columnwidth]{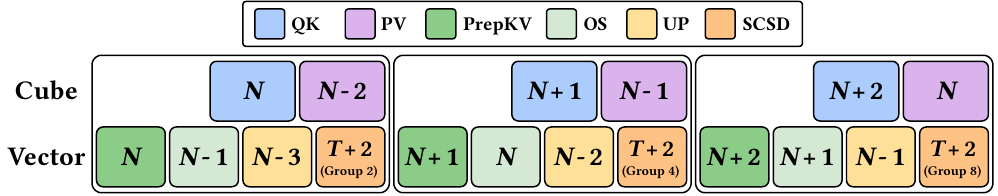}
        \caption{Illustration of pipelined SCSD execution within the attention pipeline.}
        \label{fig:design:pipeline}
    \vspace{-2mm}
\end{figure}

Although SCSD converts irregular set operations into hardware-friendly vector operations, executing it as a standalone preprocessing stage still places its latency directly on the critical path. In a straightforward implementation, the shared and query-specific Top-$K$ segments are fully constructed before the attention computation starts. The resulting execution time therefore becomes the sum of set-decomposition and attention latencies, which can substantially diminish the benefit obtained from cross-query KV sharing.

We observe that the attention kernel provides an opportunity to hide this overhead. Modern accelerators typically execute matrix-intensive and vector-intensive operations on separate compute pipelines. In sparse attention, operations such as $QK$ and $PV$ are dominated by matrix computation, whereas auxiliary operations, including softmax/update and our SCSD, mainly use vector resources. Importantly, during the steady-state attention pipeline, the vector-side workload is often shorter than the concurrent matrix computation, leaving otherwise underutilized vector execution slots. Since SCSD consists primarily of sorting, shifting, comparisons, and Boolean operations, it can naturally utilize these idle vector resources.

Based on this observation, we pipeline SCSD with the attention computation instead of executing all set decomposition upfront. The key idea is to construct each sparse segment only before it is consumed by the corresponding attention stage. We first compute a small number of initial Top-$K$ segments to prime the pipeline and break the immediate producer--consumer dependency between SCSD and attention. Once the pipeline reaches the steady state, while the matrix pipeline performs $QK$ or $PV$ computation for the current sparse segment, the vector pipeline simultaneously constructs the shared or private segment required by a future attention stage. In this way, SCSD computation is shifted away from the critical path and overlapped with useful attention computation.

To further improve the overlap, we divide the construction of a Top-$K$ segment into multiple finer-grained stages. A complete set-decomposition task may be too large to fit within a single vector-side slack window. We therefore split it into several smaller stages and schedule them across consecutive attention pipeline slots. Each stage consumes only a fraction of the available vector capacity, allowing the kernel to better utilize otherwise idle vector resources without delaying the matrix pipeline. After the warmup phase, SCSD and attention thus progress as a producer--consumer pipeline: attention consumes previously prepared sparse segments while SCSD constructs future segments in parallel.

This staged pipelining largely hides the additional bookkeeping introduced by cross-query sharing. Rather than paying
$T_{\mathrm{SCSD}} + T_{\mathrm{Attn}}$ sequentially, the effective SCSD latency is overlapped with the existing attention pipeline, making the additional set-decomposition overhead largely invisible on the critical path.

\subsection{Cascaded Sharing}

\begin{figure}[!t]
	\centering
        \setlength{\abovecaptionskip}{0.1cm}
	\includegraphics[width=0.85\columnwidth]{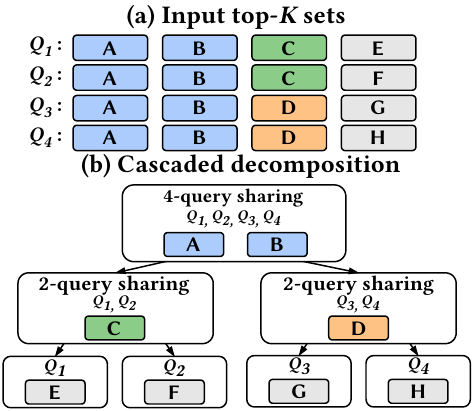}
        \caption{A running example of cascaded sharing across multiple query-group granularities.}
        \label{fig:design:cascade_sharing}
    \vspace{-5mm}
\end{figure}

Although grouping neighboring queries enables cross-query KV reuse, using a fixed sharing granularity cannot fully exploit the heterogeneous overlap structure of sparse-attention workloads. A small query group typically exhibits a high Top-$K$ overlap ratio, but misses reuse opportunities across different groups. In contrast, a large group covers a wider sharing scope, yet only KV entries selected by \emph{all} queries in the group can be directly shared, causing many entries that are shared by only a subset of queries to fall back to redundant execution.

To address this limitation, we propose \emph{cascaded sharing}, which hierarchically decomposes the sparse workload and captures KV reuse at multiple granularities. Starting from a relatively large query group (8 by default), we first extract the KV entries shared by all queries in the group and process them jointly. These shared entries are then removed from the Top-$K$ sets of individual queries, and the remaining entries are recursively partitioned into smaller subgroups to identify additional reuse opportunities. The decomposition continues until no further sharing is possible, at which point the remaining entries are processed as query-specific residuals. In this way, each KV entry is processed at the largest query granularity over which it can be shared, avoiding the need to commit to a single fixed group size.

\begin{table}[t]
\centering
\caption{Number of processed KV entries under different sharing strategies
for the running example in Figure~\ref{fig:design:cascade_sharing}.}
\label{tab:design:cascade_example}
\vspace{-2mm}
\setlength{\tabcolsep}{5pt}
\begin{tabular}{lcc}
\toprule
\textbf{Strategy} &
\textbf{Processed KVs} &
\textbf{Reduction} \\
\midrule
No sharing       & 16 & -- \\
Group = 2        & 10 & 37.5\% \\
Group = 4        & 10 & 37.5\% \\
Cascaded         & \textbf{8} & \textbf{50.0\%} \\
\bottomrule
\end{tabular}
\vspace{-4mm}
\end{table}

Figure~\ref{fig:design:cascade_sharing} illustrates cascaded sharing with four neighboring queries whose Top-$K$ sets are \(S_1=\{A,B,C,E\}\), \(S_2=\{A,B,C,F\}\), \(S_3=\{A,B,D,G\}\), and \(S_4=\{A,B,D,H\}\). Cascaded sharing first extracts the four-way intersection \(\{A,B\}\), which is loaded and dequantized once and reused by all four queries. After removing these entries, the residual workloads become \(\{C,E\}\), \(\{C,F\}\), \(\{D,G\}\), and \(\{D,H\}\), respectively. The four-query group is then divided into two subgroups, \(\{Q_1,Q_2\}\) and \(\{Q_3,Q_4\}\), where the additional shared entries \(C\) and \(D\) are extracted and reused within each subgroup. Finally, only the query-specific residuals \(E\), \(F\), \(G\), and \(H\) are processed independently. 

Table~\ref{tab:design:cascade_example} quantifies the benefit of such multi-granularity sharing in terms of the number of processed KV entries. With a fixed group size of two, the globally shared entries \(A\) and \(B\) are redundantly processed by both subgroups, resulting in 10 processed KV entries. Increasing the group size to four eliminates this inter-group redundancy, but fails to capture the subgroup-level sharing of \(C\) and \(D\), again resulting in 10 processed KV entries. In contrast, cascaded sharing captures both four-way and two-way reuse and reduces the total number of processed KV entries to 8, equal to the number of distinct KV entries in this example.

This hierarchical organization therefore avoids committing to a single fixed sharing granularity and allows each KV entry to be processed at the largest query granularity over which it can be shared.

\subsection{Tile-Aware Clipping}

\begin{figure}[!t]
	\centering
        \setlength{\abovecaptionskip}{0.1cm}
	\includegraphics[width=1\columnwidth]{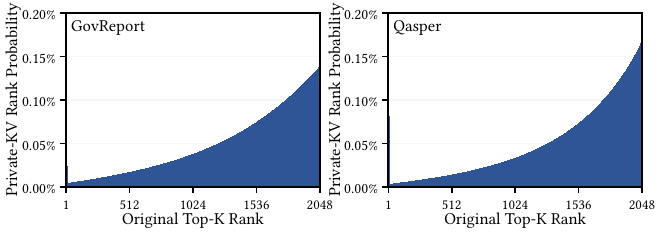}
        \caption{The ranking distributions of query-specific KVs.}
        \label{fig:design:sharing}
    \vspace{-2mm}
\end{figure}

Cascaded sharing partitions the Top-$K$ workload into multiple shared and query-specific segments whose sizes are generally not aligned with the native attention tile size. Consequently, even a few remaining KV entries may trigger an additional QK/PV tile and lead to poor hardware utilization. We therefore make cascaded sharing aware of the underlying tile granularity.

Let $T$ denote the number of KV entries per tile and $|S|=qT+r$ ($0\le r<T$) for a shared segment $S$. We introduce a tile-utilization threshold $\tau$ (50\% by default). If $r\ge\tau T$, we pad the tail to a full tile and process it at the current sharing level, masking the padded entries to preserve exact semantics. Otherwise, we push the tail down to the next cascade level, where it is merged with lower-level workloads to form better-utilized tiles. This push-down is exact and trades a small amount of sharing for higher tile utilization.

At the final query-specific level, no further push-down is possible. Figure~\ref{fig:design:sharing} shows the rank distribution of these query-specific KV entries after cascaded sharing. We observe that the remaining residual entries are predominantly concentrated near the tail of the original Top-$K$ ranking, suggesting that highly ranked KV entries are more likely to have already been captured by shared segments at earlier cascade levels. Based on this observation, we perform \emph{importance-aware clipping}: for $|R_i|=q_iT+r_i$, we retain the $q_iT$ KV entries with the highest indexer scores and prune the $r_i$ least important ones, thereby eliminating the final underutilized tile. This is the only approximate step in our tile-aware execution and minimizes accuracy degradation by discarding only low-importance residual entries.

\section{Performance Evaluation}

\subsection{Experimental Setup}
\stitle{Testbed}
We conduct our experiments on 16 Ascend 910C NPUs, each equipped with 64~GB of HBM. The NPUs are interconnected through a high-speed scale-up fabric, providing up to 784~GB/s bidirectional D2D bandwidth. We use CANN 9.1.0 as the underlying software stack for kernel development and execution. Although our evaluation is conducted on Ascend NPUs, our design does not rely on Ascend-specific architectural features. Instead, it builds on general characteristics of modern AI accelerators, including separate matrix and general-purpose/vector compute units, tile-based execution, and heterogeneous compute pipelines. Therefore, the proposed techniques can also be applied to GPUs, where similar functionality is provided by Tensor Cores and general-purpose SIMT cores.

\stitle{Models and Workloads}
We evaluate our design using two representative long-context LLMs, DeepSeek-3.2~\cite{DeepSeekDSA} and GLM-5.3~\cite{GLM5}, both of which employ model-native sparse attention. DeepSeek-3.2 uses a trainable Lightning Indexer in each attention layer to select the Top-2048 KV entries independently for each query. GLM-5.3 adopts a similar Top-2048 DSA mechanism, while further allowing neighboring layers to share indexer selections. For workloads, we use datasets from LongBench~\cite{longbench}, covering a diverse set of long-context tasks, which includes \textit{Question Answering}, \textit{Document Summarization}, and \textit{Code Generation}. These workloads cover diverse input structures and attention behaviors, allowing us to evaluate the effectiveness of our design across different long-context scenarios.

\stitle{Baseline and implementation}
We compare \name{} against the optimized sparse-attention kernel implementation provided by OPS-Transformer~\cite{opstransformer}, which represents a highly optimized implementation for Ascend NPUs. To ensure a fair comparison, we integrate both the OPS-Transformer baseline kernel and our workload-aware sparse-attention kernel into the same XYServe~\cite{xyserve} inference engine. Both implementations therefore share the same model execution framework, request scheduling, memory management, and runtime environment, with the sparse-attention kernel being the only major difference. Unless otherwise specified, all reported kernel-level and end-to-end results are measured under this unified serving setup.

\begin{table}[t]
\centering
\caption{Accuracy comparison of sparse-attention kernels across 21 LongBench datasets (\%).}
\label{tab:performance}
\vspace{-2mm}

\setlength{\tabcolsep}{3.2pt}
\renewcommand{\arraystretch}{1.08}
\footnotesize

\begin{tabular}{lccc@{\hspace{3pt}}ccc}
\toprule
\multirow{2}{*}{\textbf{Dataset}} &
\multicolumn{3}{c}{\textbf{GLM-5.3}} &
\multicolumn{3}{c}{\textbf{DeepSeek-3.2}} \\
\cmidrule(lr){2-4}
\cmidrule(lr){5-7}
&
\textbf{Baseline} &
\textbf{\name{}} &
\textbf{$\Delta$} &
\textbf{Baseline} &
\textbf{\name{}} &
\textbf{$\Delta$} \\
\midrule
qmsum                   & 22.48 & 22.60 & +0.12 & 22.86 & 23.02 & +0.16 \\
gov\_report             & 30.40 & 30.60 & +0.20 & 29.85 & 29.60 & -0.25 \\
narrativeqa             & 32.20 & 31.36 & -0.84 & 31.94 & 31.76 & -0.18 \\
samsum                  & 45.81 & 46.09 & +0.28 & 43.94 & 44.69 & +0.75 \\
dureader                & 28.64 & 28.89 & +0.25 & 26.11 & 26.84 & +0.73 \\
lsht                    & 57.50 & 57.50 &  0.00 & 47.00 & 47.75 & +0.75 \\
trec                    & 78.50 & 78.00 & -0.50 & 58.88 & 58.25 & -0.63 \\
2wikimqa                & 60.11 & 59.87 & -0.24 & 66.43 & 67.57 & +1.14 \\
hotpotqa                & 63.76 & 64.18 & +0.42 & 67.99 & 66.54 & -1.45 \\
passage\_retrieval\_zh &100.00 &100.00 &  0.00 &100.00 &100.00 &  0.00 \\
qasper                  & 46.39 & 47.05 & +0.66 & 49.57 & 49.80 & +0.23 \\
vcsum                   & 12.30 & 12.19 & -0.11 & 14.66 & 14.44 & -0.22 \\
multifieldqa\_zh        & 62.11 & 62.80 & +0.69 & 65.28 & 66.32 & +1.04 \\
triviaqa                & 91.07 & 91.60 & +0.53 & 93.98 & 92.63 & -1.35 \\
passage\_count          & 29.50 & 30.08 & +0.58 &  7.50 &  8.50 & +1.00 \\
multifieldqa\_en        & 50.45 & 51.45 & +1.00 & 55.80 & 56.14 & +0.34 \\
lcc                     & 74.35 & 73.44 & -0.91 & 73.61 & 74.09 & +0.48 \\
musique                 & 51.40 & 50.55 & -0.85 & 51.35 & 54.06 & +2.71 \\
repobench-p             & 69.69 & 70.22 & +0.53 & 42.54 & 41.39 & -1.15 \\
passage\_retrieval\_en &100.00 &100.00 &  0.00 & 98.50 & 99.50 & +1.00 \\
multi\_news             & 26.84 & 26.74 & -0.10 & 23.38 & 23.26 & -0.12 \\
\bottomrule
\end{tabular}
\vspace{-4mm}
\end{table}

\subsection{Main Results}
\subsubsection{Accuracy Retainment}\hfill


\noindent We first evaluate the accuracy impact of our workload-aware sparse-attention execution. We compare XYServe using the original OPS-Transformer sparse kernel with XYServe using \name{} across 21 LongBench datasets, following the official evaluation metric for each dataset. The results are shown in Table~\ref{tab:performance}. 
For both GLM-5.3 and DeepSeek-3.2, our kernel achieves nearly identical accuracy to the baseline, with only negligible differences across individual tasks and in the average score. Specifically, across all datasets, the mean absolute score difference between \name{} and the baseline is only 0.75 points for DeepSeek-3.2 and 0.42 points for GLM-5.3. Relative to the baseline scores, the corresponding mean absolute percentage differences are 2.01\% and 0.90\%, respectively.
These results confirm that our exact execution optimizations preserve the original sparse attention semantics, while the importance-aware clipping in our tile-aware execution introduces only negligible accuracy variance.

\subsubsection{Average Kernel Latency Reduction}\hfill

\begin{figure}[!t]
	\centering
	\includegraphics[width=1\columnwidth]{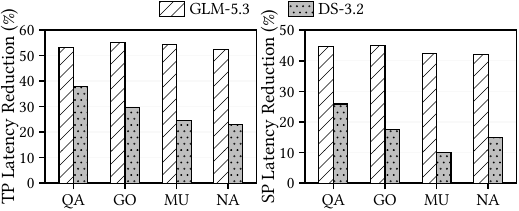}
	\caption{Average sparse-attention kernel latency reduction achieved by \name{} over OPS-Transformer on Qasper (QA), GovReport (GO), Musique (MU), and NarrativeQA (NA) from LongBench under SP and TP.}
	\label{fig:eval:e2e_kernel_latency}
	\vspace{-4mm}
\end{figure}

\begin{table}[t]
\centering
\caption{Reduction in processed KV entries achieved by \name{} over OPS-Transformer on Qasper (QA), GovReport (GO), Musique (MU), and NarrativeQA (NA).}
\label{tab:eval:processed_kv}
\vspace{-2mm}

\setlength{\tabcolsep}{8pt}
\renewcommand{\arraystretch}{1.10}
\footnotesize

\begin{tabular}{ccc}
\toprule
\textbf{Workload} &
\textbf{GLM-5.3} &
\textbf{DeepSeek-3.2} \\
\midrule
QA & 53.8\% & 39.6\% \\
GO & 55.9\% & 31.9\% \\
MU & 54.6\% & 25.2\% \\
NA & 54.3\% & 24.8\% \\
\bottomrule
\end{tabular}
\vspace{-4mm}
\end{table}

\noindent We next evaluate the kernel-level execution efficiency of \name{} on GLM-5.3 and DeepSeek-3.2. We compare \name{} against the OPS-Transformer sparse kernel on four representative LongBench workloads, including Qasper (QA)~\cite{qasper}, GovReport (GO)~\cite{govreport}, Musique (MU)~\cite{musique}, and NarrativeQA (NA)~\cite{narrativeqa}, under both tensor parallelism (TP) and sequence parallelism (SP). Figure~\ref{fig:eval:e2e_kernel_latency} reports the reduction in average prefill-stage sparse-attention kernel latency achieved by \name{} relative to OPS-Transformer.
\name{} consistently reduces kernel latency across both models, all four workloads, and both parallel configurations.
Specifically, for GLM-5.3, under TP, our kernel reduces the average sparse-attention latency by 53.2\%, 55.1\%, 54.5\%, and 52.4\% on the four datasets, respectively, corresponding to an average reduction of 53.8\%. Under SP, the reductions are 44.6\%, 45.0\%, 42.3\%, and 41.9\%, with an average reduction of 43.4\%. For DeepSeek-3.2, our kernel achieves latency reductions of 37.9\%, 29.6\%, 24.5\%, and 22.8\% under TP and 25.9\%, 17.5\%, 9.9\%, and 15.0\% under SP, corresponding to average reductions of 28.7\% and 17.1\%, respectively.

The reductions in kernel latency are accompanied by substantial decreases in KV processing.
Table~\ref{tab:eval:processed_kv} reports the reduction in processed KV
entries for each workload and model. Since TP and SP only change how the query
workload is partitioned across devices, they do not affect the total number of
KV entries processed by \name{}.
Across the four workloads, \name{} reduces the number of processed KV entries by
54.7\% on average for GLM-5.3 and 30.4\% for DeepSeek-3.2.
These results demonstrate that our design provides robust kernel-level improvements across different model architectures, workloads, and parallelization strategies by exploiting cross-query KV reuse and improving hardware utilization.

\subsubsection{Performance Sensitivity to Sharing Length}\hfill

\begin{figure}[!t]
	\centering
	\includegraphics[width=1\columnwidth]{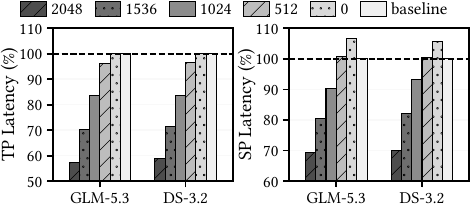}
	\caption{Average sparse-attention kernel latency with varying sharing lengths under SP and TP, normalized to OPS-Transformer.}
	\label{fig:eval:kernel_varying_sharing}
	\vspace{-2mm}
\end{figure}

\begin{figure}[!t]
	\centering
	\includegraphics[width=1\columnwidth]{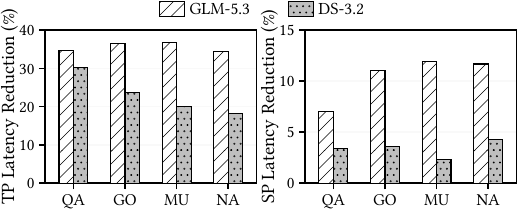}
	\caption{Average time-to-first-token latency (TTFT) reduction achieved by \name{} over OPS-Transformer on Qasper (QA), GovReport (GO), Musique (MU), and NarrativeQA (NA) from LongBench under SP and TP.}
	\label{fig:eval:e2e_model_latency}
	\vspace{-4mm}
\end{figure}

\noindent We further quantify the relationship between cross-query sharing ratio and the kernel performance using synthetic sparse-attention workloads. Specifically, we synthetically construct Top-$K$ index lists with controlled sharing lengths among neighboring queries while keeping the overall Top-$K$ size unchanged. This allows us to isolate the impact of cross-query sharing from other workload characteristics. We vary the sharing length and measure the average sparse-attention kernel latency of \name{} and OPS-Transformer for both GLM-5.3 and DeepSeek-3.2 under tensor parallelism (TP) and sequence parallelism (SP).
Figure~\ref{fig:eval:kernel_varying_sharing} shows that the kernel latency consistently decreases as the sharing length increases across both models and parallel configurations. With more shared KV entries, our workload-aware execution can eliminate more redundant KV processing and reuse the corresponding data across neighboring queries. Specifically, increasing the sharing length from 0 to 2048 reduces the kernel latency by 42.74\% and 33.25\% for GLM-5.3 under TP and SP, respectively, and by 41.03\% and 29.53\% for DeepSeek-3.2.
We also observe a small latency overhead when cross-query sharing is limited, mainly due to the set-decomposition work performed before the attention pipeline is fully overlapped. This overhead is more visible under SP because sequence partitioning shortens the per-kernel sequence length, leaving less computation to amortize the pipeline warm-up cost. As sharing increases, the reuse benefits quickly dominate this fixed overhead.
These results confirm that the performance gains of \name{} increase with the amount of cross-query sharing, as greater reuse progressively outweighs the fixed set-processing overhead.

\subsubsection{Average TTFT Reduction}\hfill

\noindent We further evaluate whether the kernel-level improvements of \name{}
translate into end-to-end prefill speedups on GLM-5.3 and DeepSeek-3.2. We
compare \name{} with OPS-Transformer in XYServe on the same four LongBench
workloads under both TP and SP, and report the reduction in average time-to-first-token latency (TTFT) relative to OPS-Transformer in Figure~\ref{fig:eval:e2e_model_latency}.
\name{} consistently reduces TTFT across both models, all evaluated workloads, and both parallel configurations. For GLM-5.3, the latency reductions under TP are 34.6\%, 36.5\%, 36.8\%, and 34.5\%, corresponding to an average reduction of 35.6\%. Under SP, the corresponding reductions are 7.0\%, 11.0\%, 11.9\%, and 11.7\%, averaging 10.4\%. DeepSeek-3.2 exhibits a similar trend, with reductions of 30.2\%, 23.7\%, 20.0\%, and 18.1\% under TP, and 3.4\%, 3.6\%, 2.4\%, and 4.3\% under SP, corresponding to average reductions of 23.0\% and 3.4\%, respectively.
These end-to-end results confirm that the benefits of our workload-aware kernel generalize across different sparse-attention models and translate into substantial reductions in user-visible prefill latency.

\begin{figure}
    \centering
    \includegraphics[width=1\linewidth]{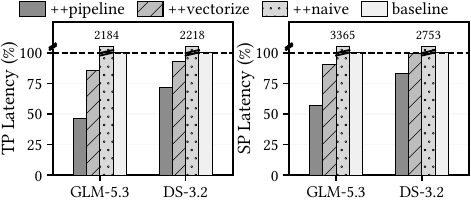}
    \caption{Average sparse-attention kernel latency under TP and SP, normalized to OPS-Transformer. Naive uses a Torch-based set-intersection implementation; Vectorized replaces it with SCSD; Pipeline further enables pipelined SCSD execution.}
    \label{fig:eval:ablations_scsd}
    \vspace{-5mm}
\end{figure}

\begin{figure}
    \centering
    \includegraphics[width=1\linewidth]{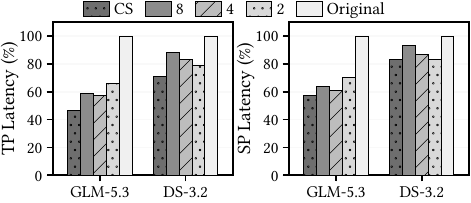}
    \caption{Average sparse-attention kernel latency of cascaded sharing (CS) and fixed query-group sizes of 2, 4, and 8 under TP and SP, normalized to OPS-Transformer. }
    \label{fig:eval:ablations_cascade}
    \vspace{-5mm}
\end{figure}

\begin{table}[t]
\centering
\caption{Average number of processed KV entries with fixed query-group sizes
and cascaded sharing (CS), normalized to OPS-Transformer.}
\label{tab:eval:cascade_kv}
\vspace{-2mm}

\setlength{\tabcolsep}{8pt}
\renewcommand{\arraystretch}{1.10}
\footnotesize

\begin{tabular}{ccc}
\toprule
\textbf{Strategy} &
\textbf{GLM-5.3} &
\textbf{DeepSeek-3.2} \\
\midrule

Group 2 & 65.7\% & 78.9\% \\
Group 4 & 57.2\% & 82.9\% \\
Group 8 & 59.0\% & 88.4\% \\
CS      & \textbf{45.5\%} & \textbf{70.1\%} \\

\bottomrule
\end{tabular}
\vspace{-5mm}
\end{table}

\subsection{Ablation Studies}
We conduct a series of ablation studies on requests from the same four LongBench workloads, namely Qasper (QA), GovReport (GO), Musique (MU), and NarrativeQA (NA), to quantify the individual contributions of the key optimizations in \name{}, including Shift-and-Compare Set Decomposition (SCSD), Pipelined Execution, Cascaded Sharing, and Tile-Aware Clipping.


\subsubsection{Effectiveness of SCSD and Pipelined Execution}\hfill

\noindent We first evaluate the effectiveness of SCSD and pipelined execution by incrementally enabling them on top of the OPS-Transformer baseline. 
We compare four variants: (1) the original baseline kernel without cross-query sharing; (2) Naive, which enables group-based query execution but computes set intersections using a naive Torch-based implementation; (3) Vectorized, which replaces the naive set operations with the proposed SCSD; and (4) Pipeline, the full design that further overlaps SCSD with attention computation. We evaluate all variants on GLM-5.3 and DeepSeek-3.2 across the same four LongBench workloads under both TP and SP. The results are shown in Figure~\ref{fig:eval:ablations_scsd}.

The naive implementation incurs substantial overhead across both models and parallel configurations. For GLM-5.3, it is 21.84$\times$ and 33.65$\times$ slower than OPS-Transformer under TP and SP, respectively. Replacing the naive set operations with SCSD reduces the kernel latency by 96.1\% under TP and 97.3\% under SP, while further enabling pipelined execution provides additional reductions of 45.9\% and 36.9\%, respectively. DeepSeek-3.2 exhibits a similar trend. The naive implementation is 22.18$\times$ and 27.53$\times$ slower than the baseline under TP and SP, respectively; SCSD reduces the latency by 95.8\% and 96.4\%, and pipelined execution further reduces it by 22.9\% and 16.2\%, respectively.
Overall, the full design outperforms the baseline by 35.5\% on average across both models and both parallel configurations. The consistent improvements demonstrate that efficient set decomposition and pipeline overlap are both necessary to translate cross-query sharing into actual performance gains.

\subsubsection{Effectiveness of Cascaded Sharing}\hfill

\noindent We next evaluate the effectiveness of cascaded sharing by comparing it with fixed query-group sizes of 2, 4, and 8. All variants use the same SCSD and pipelined execution and differ only in how cross-query sharing is organized. We evaluate GLM-5.3 and DeepSeek-3.2 on the same four LongBench workloads under both TP and SP. Figure~\ref{fig:eval:ablations_cascade} reports the normalized sparse-attention kernel latency, while Table~\ref{tab:eval:cascade_kv} reports the corresponding number of processed KV entries, also normalized to OPS-Transformer.



For GLM-5.3, cascaded sharing reduces the number of processed KV entries by
30.8\%, 20.6\%, and 22.9\% compared with fixed group sizes of 2, 4, and 8,
respectively. These reductions translate into kernel-latency improvements of
29.6\%, 19.2\%, and 21.6\% under TP, and 18.8\%, 6.8\%, and 10.9\%
under SP. DeepSeek-3.2 shows a similar trend: cascaded sharing reduces
processed KV entries by 11.2\%, 15.4\%, and 20.7\% over group sizes of 2, 4, and 8,
respectively, while reducing kernel latency by 9.7\%, 14.0\%, and 19.4\%
under TP and by 0.5\%, 4.5\%, and 11.0\% under SP.
Cascaded sharing captures both by hierarchically exploiting reuse at multiple granularities, converting more cross-query Top-$K$ overlap into reduced data movement and lower execution latency.

\subsubsection{Effectiveness of Tile-Aware Clipping}\hfill

\begin{figure}
    \centering
    \includegraphics[width=1\linewidth]{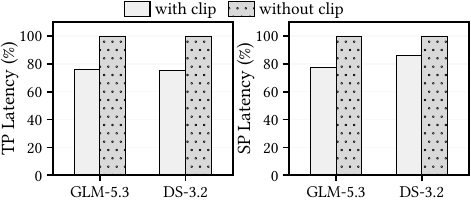}
    \caption{Average sparse-attention kernel latency with tile-aware clipping under TP and SP, normalized to the corresponding execution without clipping.}
    \label{fig:eval:ablation_clipping_latency}
    \vspace{-4mm}
\end{figure}

\begin{figure}
    \centering
    \includegraphics[width=1\linewidth]{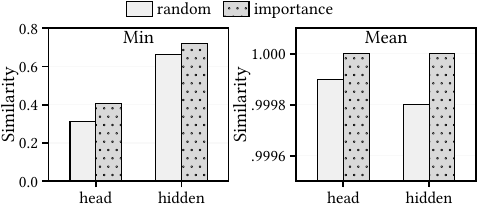}
    \caption{Cosine similarity between the attention output tensors produced by random and importance-aware clipping and the corresponding unclipped reference.}
    \label{fig:eval:ablation_clipping_similarity}
    \vspace{-4mm}
\end{figure}

\stitle{Latency Comparison}
We finally evaluate tile-aware clipping from both performance and numerical-fidelity perspectives. For performance, we compare the full \name{} design against an exact variant that retains all query-specific residual KV entries and therefore performs no clipping. Both variants otherwise use the same SCSD, cascaded sharing, and pipelined execution. As shown in Figure~\ref{fig:eval:ablation_clipping_latency}, tile-aware clipping consistently reduces sparse-attention kernel latency under both TP and SP. For GLM-5.3, the latency reductions are 23.71\% under TP and 22.68\% under SP, while DeepSeek-3.2 achieves reductions of 24.44\% and 14.13\%, respectively. These results demonstrate that eliminating poorly utilized tail tiles provides consistent performance benefits across both models and parallel configurations.

\stitle{Output Similarity Comparison}
We next examine whether importance-aware clipping preserves the attention output more effectively than arbitrary pruning with GLM-5.3. Using the unclipped execution as the reference, we compare two strategies that remove the same number of KV entries from the query-specific residuals: \emph{random clipping}, which removes residual entries uniformly at random, and our \emph{importance-aware clipping}, which removes entries with the lowest indexer scores. We measure the cosine similarity between the clipped and unclipped attention outputs as

\[
\mathrm{Sim}(O,\hat{O}) =
\frac{\langle O,\hat{O}\rangle}
     {\|O\|_2\|\hat{O}\|_2}.
\]


where $O$ and $\hat{O}$ denote the corresponding unclipped and clipped output vectors, respectively. We evaluate this similarity at two granularities. For the \emph{head-level} metric, cosine similarity is computed independently for each $D$-dimensional attention-head output. For the \emph{hidden-level} metric, all $N$ heads are flattened into a single $ND$-dimensional vector before computing cosine similarity. For both metrics, we report the mean and minimum similarity across the evaluated outputs, capturing the overall numerical fidelity and the worst-case deviation, respectively.


As shown in Figure~\ref{fig:eval:ablation_clipping_similarity}, importance-aware clipping consistently preserves the attention output more accurately than random clipping. Specifically, importance-aware clipping achieves a head-level mean cosine similarity of 1.0000, compared with 0.9999 for random clipping, while improving the head-level minimum similarity from 0.312 to 0.406. At the hidden level, the mean similarity increases from 0.9998 to 1.0000, and the minimum similarity improves from 0.664 to 0.719. These results show that pruning residual KV entries according to their indexer scores substantially reduces the numerical perturbation caused by clipping, allowing \name{} to eliminate underutilized tail tiles while closely preserving the unclipped attention output.

\section{Related Work}
\stitle{Sparse attention for prefill}
A growing body of work exploits attention sparsity to accelerate the prefill phase of long-context LLM inference~\cite{mInference, FlexPrefill, XAttention, SeerAttention, DeepSeekDSA, FlashPrefill}. MInference~\cite{mInference} identifies several recurring sparse attention patterns and dynamically constructs sparse indices for efficient prefill computation. FlexPrefill~\cite{FlexPrefill} further adapts the sparse pattern and computation budget to individual inputs and attention heads, while XAttention~\cite{XAttention} employs lightweight scoring to identify important attention blocks. SeerAttention~\cite{SeerAttention} replaces hand-crafted sparsity patterns with a learnable mechanism for predicting block-level attention sparsity. 
Different from these post-hoc or inference-time sparsification approaches, DeepSeek Sparse Attention (DSA)~\cite{DeepSeekDSA} incorporates fine-grained sparse attention directly into model training and employs a lightweight indexer to dynamically select important KV entries during both training and inference. Following DSA, model-native sparse attention has been increasingly adopted and extended by recent LLMs, including the GLM series~\cite{GLM5}, Hunyuan series~\cite{hunyuan4}, and DeepSeek-4~\cite{deepseekv4, deepseek41}. 
Meanwhile, systems such as FlashPrefill~\cite{FlashPrefill} further improve sparse prefill through hardware-aware pattern discovery and execution. 
These approaches focus on determining \emph{which} KV entries should participate in attention or on efficiently executing the sparse workload of individual queries. In contrast, \name{} operates on the resulting query-specific sparse KV sets and exploits sharing across neighboring queries, which is orthogonal to these sparsification techniques, and can be applied on top of their designs to further reduce redundant cross-query KV accesses and computation.

\stitle{Sparse attention for decoding}
A large body of work exploits attention sparsity to reduce the KV-cache access cost during autoregressive decoding. H$_2$O~\cite{h2o}, StreamingLLM~\cite{streamingllm}, Scissorhands~\cite{scissorhands}, FastGen~\cite{fastgen}, and SnapKV~\cite{snapkv} compress the KV cache by retaining only tokens considered important for future attention computation, thereby reducing both memory consumption and attention cost. To avoid permanently discarding potentially useful KV entries, dynamic sparse-attention methods such as Quest~\cite{quest}, InfLLM~\cite{infllm}, and ArkVale~\cite{arkvale} dynamically select query-relevant KV entries at runtime while retaining or recovering the full historical context. Several systems further exploit the temporal locality of sparse KV accesses across consecutive decoding steps. RetroInfer~\cite{retroinfer}, LiteCache~\cite{litecache}, and SparseServe~\cite{sparseserve} leverage the high similarity of KV selections between neighboring decoding queries to cache frequently accessed KV entries in HBM, thereby reducing costly KV transfers between host DRAM and HBM. These approaches primarily exploit cross-query locality at the memory-management level during decoding. In contrast, our \name{} targets concurrently processed queries during prefill and exploits their overlapping sparse workloads inside the attention kernel, eliminating redundant KV movement across the HBM--on-chip-memory boundary as well as redundant dequantization and computation.

\stitle{Hardware-efficient attention kernels}
A rich line of work optimizes attention kernels by better exploiting the memory hierarchy and parallel compute resources of modern accelerators~\cite{flashattention, flashattention2, flashattention3, flashinfer}. FlashAttention~\cite{flashattention} introduces an IO-aware tiled attention algorithm that keeps intermediate attention states in on-chip memory and employs online softmax to avoid materializing the full attention matrix in HBM, substantially reducing HBM accesses. FlashAttention-2~\cite{flashattention2} further improves hardware utilization through better work partitioning across thread blocks and warps, increased parallelism, and reduced non-matrix-multiplication overhead. FlashAttention-3~\cite{flashattention3} exploits emerging GPU features such as asynchronous matrix computation and data movement, using warp specialization and pipelining to overlap memory accesses, matrix multiplication, and softmax computation, while also supporting low-precision execution. FlashInfer~\cite{flashinfer} provides optimized and customizable attention kernels for diverse LLM inference workloads and KV-cache layouts. These works establish general principles for efficient attention execution, including maximizing on-chip data reuse, improving compute-unit utilization, and overlapping computation with data movement. Our \name{} builds upon these principles but targets a complementary source of inefficiency specific to sparse attention: it exploits the sharing structure of Top-K KV selections across neighboring queries to expose additional cross-query data reuse that conventional attention kernels do not capture.

\section{Conclusion}
In this work, we revisit sparse-attention execution for long-context LLM prefill and identify substantial cross-query KV reuse that is largely overlooked by existing workload-agnostic, query-parallel execution. 
Based on this observation, we propose \name{}, a workload-aware sparse-attention execution mechanism that jointly processes correlated queries to exploit cross-query sharing. \name{} introduces Shift-and-Compare Set Decomposition (SCSD) for efficient data-parallel set processing, pipelines SCSD with attention computation to hide its overhead, and employs cascaded sharing to capture reuse across multiple query granularities. It further incorporates tile-aware execution to improve hardware utilization, together with importance-aware clipping to retain high accuracy.
Evaluations on GLM-5.3 and DeepSeek-3.2 show that \name{} consistently reduces sparse-attention kernel latency and translates these improvements into substantial end-to-end prefill speedups across diverse long-context workloads and parallel configurations. Overall, our results demonstrate the importance of exploiting cross-query workload structure, rather than treating sparse queries independently, for efficient long-context LLM inference.

\bibliographystyle{ACM-Reference-Format}
\bibliography{sample-base}

\end{document}